\documentclass{article}
\usepackage{float} 
\usepackage{authblk}
\usepackage[english]{babel}
\usepackage{amsthm,amsmath,amssymb}
 
\usepackage{mathrsfs}
\usepackage[letterpaper,top=2cm,bottom=2cm,left=3cm,right=3cm,marginparwidth=1.75cm]{geometry}
\usepackage{multirow}
\usepackage{makecell}
\usepackage{amsmath}
\usepackage{amsfonts,amssymb}
\usepackage{graphicx}
\usepackage[colorlinks=true, allcolors=blue]{hyperref}

\title{Attention is all you need for boosting graph convolutional neural network}

\author[a]{Yinwei Wu \thanks{The final year project work was carried out under the 3+1+1 Educational Framework at the National University of Singapore (Chongqing) Research Institute.}}
\affil[a]{College of Software Engineering, Sichuan University, Chengdu, China}

\begin{document}

\maketitle

\begin{abstract}
Graph Convolutional Neural Networks (GCNs) possess strong capabilities for processing graph data in non-grid domains. They can capture the topological logical structure and node features in graphs and integrate them into nodes' final representations. GCNs have been extensively studied in various fields, such as recommendation systems, social networks, and protein molecular structures. With the increasing application of graph neural networks, research has focused on improving their performance while compressing their size. In this work,  a plug-in module named Graph Knowledge Enhancement and Distillation Module (GKEDM) is proposed. GKEDM can enhance node representations and improve the performance of  GCNs by extracting and aggregating graph information via multi-head attention mechanism. Furthermore, GKEDM can serve as an auxiliary transferor for knowledge distillation. With a specially designed attention distillation method, GKEDM can distill the knowledge of large teacher models into high-performance and compact student models. Experiments on multiple datasets demonstrate that GKEDM can significantly improve the performance of various GCNs with minimal overhead. Furthermore, it can efficiently transfer distilled knowledge from large teacher networks to small student networks via attention distillation.
\end{abstract}

\newpage
\section*{Acknowledgement}
This work was carried out under the 3+1+1 Educational Framework at the National University of Singapore(Chongqing) Research Insitute. The author is honored to have the support of Prof. Wang Xinchao from the National University of Singapore, Mr. Wang Xu from Sichuan University and Mr. Jing Yongcheng from University of Sydney.

\newpage
\tableofcontents
\newpage
\listoffigures
\newpage
\listoftables

\newpage
\section{Introduction}
Graph Convolutional Neural Networks (GCNs) are powerful representation learning tools for graph data in non-grid domains. GCNs can learn graph data structures through generalization of convolution, which is widely used in computer vision domains. Graph neural networks have made influential achievements in social networks, recommendation systems, biomedical research and other domains.
Message-passing based GCNs are the prevailing architecture currently. This type of GCN leverages a message passing algorithm on the graph to capture node interaction and aggregates node features hierarchically to extract global features. Through message-passing, nodes obtain information about themselves and neighbors to update their representations. Message-passing algorithms typically involve neighbor aggregation, non-linear activations, and techniques like convolution, pooling, and residual for complex feature transformations and model optimization. GCNs based on message propagation mechanism stack multiple graph convolutional layers to achieve a larger receptive field, enabling better topology capturing of the graph.
However, One major challenge hindering the advancement of GCNs is the issue of over-smoothing. This refers to a phenomenon in which node representation in a multi-layer GCN gradually becomes indistinct or overly smoothed as they pass through layers, making it challenging to distinguish between them. The root cause of this problem is that in each layer, representations are updated by averaging or weighted averaging with neighboring nodes' representations, which can result in a loss or muddling of information and cause node features to become increasingly similar. This limitation hinders GCNs from adding extra layers and therefore restricts their potential to become more expressive models.

with the deepening of the research, the problem of over-smoothing has gradually been alleviated \cite{Li_Müller_Ghanem_Koltun_2021}\cite{Ying_Cai_Luo_Zheng_Ke_He_Shen_Liu_2021}. As the over-smoothing problem was alleviated, researchers began gradually increasing the number of parameters of the GCNs in order to boost their expressive capabilities, just as in the image and text domains. This trend makes it a challenge to deploy graph neural networks under resource and time constraints.


Knowledge distillation was proposed by \cite{Hinton_Vinyals_Dean_2015} in 2015, and has now become the mainstream method of model compression. Knowledge distillation provides additional supervision signals for the training of the student network to improve its performance. The additional supervision signals come from a trained teacher model with a large number of parameters, which extracted the hidden distribution of the complex training data. Initially, knowledge distillation relied solely on soft target probabilities imparted by teachers to facilitate the learning of student networks. Recently, to further investigate the potential of knowledge distillation, many studies are no longer confined to the soft target probabilities. Instead, they also utilize the intermediate layers of the model for distillation \cite{Tung_Mori_2019}\cite{Park_Kim_Lu_Cho_2020}.

Knowledge distillation in traditional neural networks for image and text data has been extensively studied. However, research on distilling knowledge in graph neural networks is still in its infancy \cite{Yang_Qiu_Song_Tao_Wang_2020}\cite{Joshi_Liu_Xun_Lin_Foo_2022}\cite{He_Wang_Zhang_Wu_2022}.
Graph data contains unique information, such as its topological structure, and is distinct from grid data. To improve distillation efficiency, careful selection and design of the information to be distilled is necessary.

In this work, a novel plug-in module for GCNs named GKEDM (Graph Knowledge Enchance and Distillation Module), is proposed. GKEDM is capable of replacing the output layer of GCN. Its knowledge enhancement module employs an attention mechanism for local topology information aggregation to boost node representation, thus improving network performance.

Simultaneously, to alleviate the computational burden resulting from  parameters, a custom attention graph knowledge distillation method that suitable for GKEDM modules is introduced. This method enables the teacher network to transfer local topology to the student model, achieving the goal of distillation.

To demonstrate the universality of GKEDM, I performed enhancement and knowledge distillation on different kinds of GCN. Experimental results across multiple datasets demonstrate that GKEDM significantly enhances the capabilities of graph neural networks and optimizes the efficiency of knowledge distillation.

The contributions of this paper are:
\begin{itemize}
\item A plug-in GCN enhancement module based on graph local topology attention is proposed, named GKEDM. It can be directly applied to GCNs and improve its performance.

\item A novel knowledge distillation method suitable for GKEDM is introduced, which can effectively transfer the topology knowledge of GCN from teacher network to student network.

\item The experimental results show that the GKEDM is versatile and effective in various GCNs and datasets.
\end{itemize}




\newpage
\section{Related work}
This work is related to graph convolutional neural networks, knowledge distillation on graph convolutional neural networks, and attention mechanisms.
\subsection{Graph convolutional neural network}
In recent years, the domain of graph neural networks(GNN) has gained significant research attention owing to its powerful processing capabilities on non-grid data. After the convolution operation on the graph was defined \cite{Kipf_Welling_2016}, an increasing number of convolution-based approaches are being used to enhance the representational capacity of GNNs. That specific type of GNN is referred to as GCN. Graph Attention Network(GAT) \cite{Liu_Zhou_2022} is an attention mechanism-based GCN designed for processing graph data. Compared to other GCNs, GAT utilizes a trainable inter-node attention mechanism to evaluate the relevance of each node to the target node. It then performs weighted aggregation of representations of neighboring nodes to better capture the local topology. his attention mechanism adapts to the representations of neighboring nodes, making GAT more flexible and comprehensible while processing graph data.
GraphSAGE \cite{Hamilton_Ying_Leskovec_2017} leverages self-supervised learning to update a node's embedding based on its neighbors, providing flexibility in handling diverse graph data types.With neighbor sampling, it can efficiently process large-scale graph data. 
GCNII \cite{Chen_Wei_Huang_Ding_Li_2020} adopts residule block to prevent over-smoothing while training, thereby enabling the stacking of more GCN layers. 
MONET \cite{Monti_Boscaini_Masci_Rodola_Svoboda_Bronstein_2017} incorporates a learnable distance-based kernel function that transforms similarity between nodes into Gaussian Kernel Functions at multiple scales. This approach retains information at various levels, resulting in improved node representations.
Rather than designing a new structure of GCN, this work focuses on improving the performance of the existing GCNs, and performing knowledge distillation and model compression on them.

\subsection{Graph Neural network knowledge distillation}
The data to be processed by convolutional neural networks(CNN) performing on grid domain data is often a fixed-size matrix, such as image data. Different from traditional CNN, the GCNs need to deal with the non-regular graph data composed of nodes and edges.
With additional features that grid domain data does not have, such as node features, graph topology, communities, etc., GCNs usually need to be carefully designed. Taking into account the aforementioned dissimilarities, the knowledge distillation approach for GCNs must also address the distillation of supplementary graph data structure captured. GCN knowledge distillation was first proposed in \cite{Yang_Qiu_Song_Tao_Wang_2020}. Yang et al. captured the realationship between nodes by introducing an LSP(Local Structure preserving) module to achieve local topology distillation in graphs. More specifically, LSP uses the distance of the node representation in the kernel space to measure the similarity between nodes, and regards the similarity of the first-order neighborhood of nodes as a distribution. By letting the student network mimic the node neighborhood distribution of the teacher network, the purpose of distillation topology knowledge can be achieved. Subsequently, other methods of graph neural network knowledge distillation were proposed. In \cite{He_Wang_Zhang_Wu_2022}, He et al. used an adversarial way to perform model compression, named GraphAKD. In GraphAKD, the student network as a generator needs to produce output similar to the teacher network to fool the discriminator, and the discriminator needs to distinguish between the output of the student and the output of the teacher.
Contrastive learning distillation has been proposed in the image domain \cite{Tian_Krishnan_Isola_2019}, and contrastive learning in the graph neural network domain has also been used as a way of representation learning \cite{Oord_Li_Vinyals_2018} to prove effective. CKJ et al. \cite{Joshi_Liu_Xun_Lin_Foo_2022} combined graph neural network distillation and contrastive learning to propose G-CRD. Through contrastive learning, G-CRD makes the representation of the student network node approximate the representation of the corresponding node of the teacher network, and at the same time, it is far away from the representation of other nodes of the teacher network, implicitly preserving the relationship between nodes, and achieving the purpose of topology distillation.
The knowledge distillation approach introduced in this research differs from previous approaches, as it focuses on a specific architecture for graph neural networks with GKEDM enhancement.

\subsection{Attention mechanism}
Attention mechanism (AM) is an important technique for enhancing the expressive ability of neutral network models. Attention allows for varying weights to be assigned to different segments of input data. This enhances neural networks' ability to manage complex data structures and multiple sources of information.
The AM is actually reflected in the very early gate mechanism, but it received extensive attention after the Transformer \cite{Vaswani_Shazeer_Parmar_Uszkoreit_Jones_Gomez_Kaiser_Polosukhin_2017} framework came out.  Compared to traditional convolution, AM has stronger relationship capturing abilities and significant potential applications in the era of big data. Transformer was first applied to the natural language processing domain, and then migrated to the image domain \cite{Dosovitskiy_2020}, which has also proved to be very effective. All the current giant models with amazing results are basically evolved based on Transformer architecture.

In the graph neural network domain, the attention mechanism was first used in GAT \cite{Liu_Zhou_2022}. Unlike Transformer, the GAT does not use the framework of Query, Key, and Value, but the idea based on attention still achieves good results. Since the GAT is implemented based on the message passing mechanism, there can be problems with over-smoothing when training large deep models.
In order to solve the oversmoothing problem and design a Transformer suitable for graph data, Ying et al. abandoned the architecture based on the message passing mechanism, and carefully designed a very important feature in the Transformer in the proposed Graphormer \cite{Ying_Cai_Luo_Zheng_Ke_He_Shen_Liu_2021}, which is positional encoding. Positional encoding has been proposed in the Transformer of the original version for natural language processing. Since Transformer does not have explicit position information, and order information is crucial in text sequences, a mechanism is needed to represent the position of each input. To solve this problem, Transformer added positional encoding to explicitly identify the position information of the input for the model. Positional encoding has been proved to have a significant influence on the results in the design of Transformer, so subsequent Transformer have also adopted a variety of positional encodings suitable for different data structures. 
Graphormer can excel in graph structures by creating appropriate positional encodings to denote the roles, locations, and other characteristics of nodes within the graph. However, Graphormer, as a derivative architecture of Transformer, also has the most important problem, that is computational complexity. And because of the particularity of the graph structure, this problem becomes more serious.Each node must query and compute with all nodes in the graph, resulting in a computational complexity of $O(n^2)$. This challenge presents a major obstacle for the application of Graphormer to graph datasets containing tens of thousands of nodes.

This work employs an attention mechanism that leverages the strengths of GAT and Graphormer and incorporates a self-attention mechanism based on local structure, thereby reducing computational load. This work incorporates a self-attention mechanism utilizing Query, Key, and Value architectures, while also implementing positional encoding based on the graph Laplacian matrix and random walk to enhance the model's expressive capability.

\newpage
\section{Background}
The purpose of this work is to enhance and distill GCNs based on attention mechanism. This section first introduces some basic principles and algorithms of graph representation learning based on message passing mechanism, and is followed by an introduction related to the self-attention mechanism and knowledge distillation.
\subsection{Notations}
Due to the large number of symbols involved, this subsection provides a comprehensive listing of all mathematical symbolic representations employed in this work, as well as their corresponding descriptions, for the sake of clarity and convenience.
The notations are shown in Table \ref{tab:notations}:

\begin{table}[h]
\centering
\begin{tabular}{c|l}
Notation & Description \\\hline
$G=(\mathcal{V}, E)$ & A graph $G$ containing the set of edges $E$ and the set of nodes $\mathcal{V}$ \\\hline
$v_n \in \mathcal{V}$ & Denotes the node $n$ \\\hline
$e_{ij} \in E$ & Denotes the edge connecting nodes $i$ to $j$ \\\hline
 \multirow{2}{*}{$H_{i}^{T}, H_{i}^{S}$} &Denotes all the node representations at the output of layer $i$ of the teacher GCN and\\
 &all the node representations at the output of layer $i$ of the student, respectively \\\hline
 $h_{n}^i$ & Denotes the representation of node $n$ at layer $i$ \\\hline
 $z_n$ & Denotes the logits of node $n$ predicted by the GCN \\\hline
 $y_{n}$ & Denotes the label of node $n$ \\\hline
 ${y_n^{pred}}$ & Denotes the prediction result of the network for node $n$ \\\hline
 $x_{n}$ & Denotes the initial features of node $n$\\\hline
$N(v_n)$ & Denotes 1-hop neighbors of node $n$\\\hline
$Q, K, V$ & Denotes $Query, Key, Value$ in attention mechanism\\\hline
$d$ & Denotes the length of the node embedding vector\\\hline
\end{tabular}
\caption{\label{tab:notations}Notations Table}

\end{table}

\subsection{Graph nerual network and Graph representation learning}
Graph neural networks (GNNs) are neural networks that can generate node embeddings based on node and edge features as well as graph topology. These embeddings can be further used for various downstream tasks.
Types of tasks that GNN can perform include node prediction, graph prediction, edge prediction, and relationship prediction. This work primarily focuses on node classification of graph convolutional neural networks (GCN). Therefore, this section delves into the GCN and its message passing mechanism, which is a fundamental algorithm for performing classification predictions.

GCN receives graph that composed of edge set $E$ and node set $\mathcal{V}$ as input. For each node $v_n \in \mathcal{V}$ there exists an initial $d-dimensional$ input feature vector $x_{n}$, and in general, this initial feature is the input to layer 0 of the GCN. Also, each node has a corresponding label $y_n$, and 1-hop neighbor node set $N(v_n)$. For the node classification task, the GCN needs to predict a label ${y_n^{pred}}$ for each node based on its feature information and the topology of the graph, and make ${y_n^{pred}}$ and $y_n$ as similar as possible in terms of metrics.
A message passing GCN consists of a stack of convolutional layers, each of which can be viewed as an encoder. In each layer of the GCN, the nodes embed the topology in their own information by collecting information from their neighbors. This stage is known as $AGGREGATE$. Then the node enters the second $UPDATE$ phase to generate and update its own node representation in the next layer. The above aggregation-update process can be expressed in mathematical language as the equation (\ref{con:equ1}):
\begin{equation}
    \begin{aligned}
        message_{N(v)}^{i} = AGGREGATE^{(i)}({h_{u}^i | \forall u \in N(v)})\\
        h_v^{i+1} = UPDATE^{(i)}(h_v{^i}, message_{N(v)}^{i}) \label{con:equ1}
    \end{aligned}
\end{equation}

In the above equation, $AGGREGATE$ is an aggregation function with permutation-invariant, $message$ is the information of the local neighborhood that the node aggregated, and $h_v^{i+1}$ is the input representation of the node in the next layer, specifically, $h_v^{0}=x_v$. After a total of $i$ layers of message passing convolutional layers are stacked, the node can aggregate the information of the nodes in the $i$-order neighborhood and thus has a more powerful representation.  For the node classification task, the GCN ultimately passes the node's final representation to a linear classifier, such as a multilayer perceptron (MLP). This step calculates the probability distribution $z_v$ for each class and determines the loss $L_{CE}$ using the cross-entropy loss function as stated in Eq. (\ref{con:equ2}):
\begin{equation}
    \begin{aligned}
        z_v &= MLP(h_v^i) \\
        L_{CE} &= H(y_v, z_v)  \label{con:equ2}
    \end{aligned}
\end{equation}

$H$ denotes the cross-entropy loss function. After the training is completed by gradient descent, the prediction results of the nodes can be obtained by $softmax(z_v)$.

\subsection{Multi-headed self-Attention mechanism}
Self-Attention mechanism based on $Key, Query, Value$ has received extensive attention after being proposed in Transformer\cite {Vaswani_Shazeer_Parmar_Uszkoreit_Jones_Gomez_Kaiser_Polosukhin_2017}. 
Its main principle is to map the set of input feature vectors $H \in \mathbb{R}^{n \times d}$ to three new sets of features located in different expression spaces , $K\in \mathbb{R}^{n \times d}, Q\in \mathbb{R}^{n \times d}, V\in \mathbb{R}^{n \times d}$, by a function as shown in Eq. (\ref{con:equ3}):

\begin{equation}
    \begin{aligned}
        K = f_k(H), \\
        Q = f_q(H), \\
        V = f_v(H), \\
    \end{aligned} \label{con:equ3}
\end{equation}
 For any two input nodes, the similarity is computed by point multiplication using the $Query$ features of the source node and the $Key$ features of the target node. Once the similarity between the target node and all the source nodes is obtained, the target node weights and sums the value features of these source nodes based on the calculated similarity to produce its own new representation. The mathematical expression is shown in the Eq. (\ref{con:equ4}):

\begin{equation}
    \begin{aligned}
        Attention(Q,K,V) = Softmax(QK^T/\sqrt{d})V,
    \end{aligned} \label{con:equ4}
\end{equation}

$QK^T$ is a $n \times n$ matrix that represents the similarity between two two nodes, and $\sqrt{d}$ is a normalization factor. The final resulting $Attention(Q,K,V)$ matrix has the same shape as $V$, so that the Transformer can improve the expressiveness of the model by stacking multiple identical attention layers.

In order to enable the model to observe more representation space, Multi-head Self Attention (MSA) is one of the main extensions to enhance the representational power of Transformer models. 
The MSA is based on the idea of splitting, where the input vector is partitioned into multiple subvectors and each subvector is assigned a separate attention head. Each attention header contains an independent set of attention weight matrices for computing the self-attention. Finally, the self-attention results of all heads are aggregated to form the final output vector.
For a MSA using $k$ heads, the model will use $K$ sets of different functions to obtain $K$ sets $Q,K,V$ and perform the self-attentive mechanism separately. Each head will provide a matrix in the shape of $V$. Finally, for all the results of self-attention, a linear mapping or averaging can be used to obtain the output. This approach can enhance the performance of the Transformer to some extent.

\subsection{Knowledge distillation}
Knowledge distillation is a technique for enhancing the performance of student models through the transfer of advanced and complex knowledge from large and high-performance teacher models. 
By feeding the output of the teacher model as a supervised signal to the student model, the student model can leverage the experience and knowledge of the teacher model to enhance its ability to learn the distinguishing features and patterns of the task, ultimately improving overall performance. 
The knowledge distillation first proposed in \cite{Hinton_Vinyals_Dean_2015} by having the student output label probabilities to match the teacher output soft label probabilities, with a specific loss function formula of Eq. (\ref{con:equ5}):
\begin{equation}
    \begin{aligned}
        L_{KD} = H(softmax(z^T/T_1), softmax(z^S/T_1))
    \end{aligned} \label{con:equ5}
\end{equation}

where $H$ denotes the cross-entropy loss or KL divergence, and $T_1$ denotes the distillation temperature, which allows for smoother labels for distillation. $z^T$ and $z^S$ are the labeling probabilities of teachers and students, respectively.



\newpage
\section{Methods}
In this section, we will briefly introduce the motivation for proposing GKEDM and present the principles and specific algorithms of GKEDM.

\subsection{Graph knowledge enhancement module}
\subsubsection{Reason}
Message passing  based GCN embeds topology into the representation vector of nodes by aggregation of multiple convolutional layers. However, overstacking the number of convolutional layers can cause over-smoothing, i.e., the representation of all nodes tends to be the same and cannot be differentiated. One of the reasons for this phenomenon is that GCNs indiscriminately aggregate information from node neighbors, and some information from important neighbors may be underestimated, while some noisy information from completely unrelated neighbors may be amplified. Over-smoothing phenomenon becomes more and more significant as the number of layers of the GCN increases, which in turn leads to degradation of the model performance.

\begin{figure}[H]
    \centering
    \label{fig:oversmoothing}
    \includegraphics[width=0.8\textwidth]{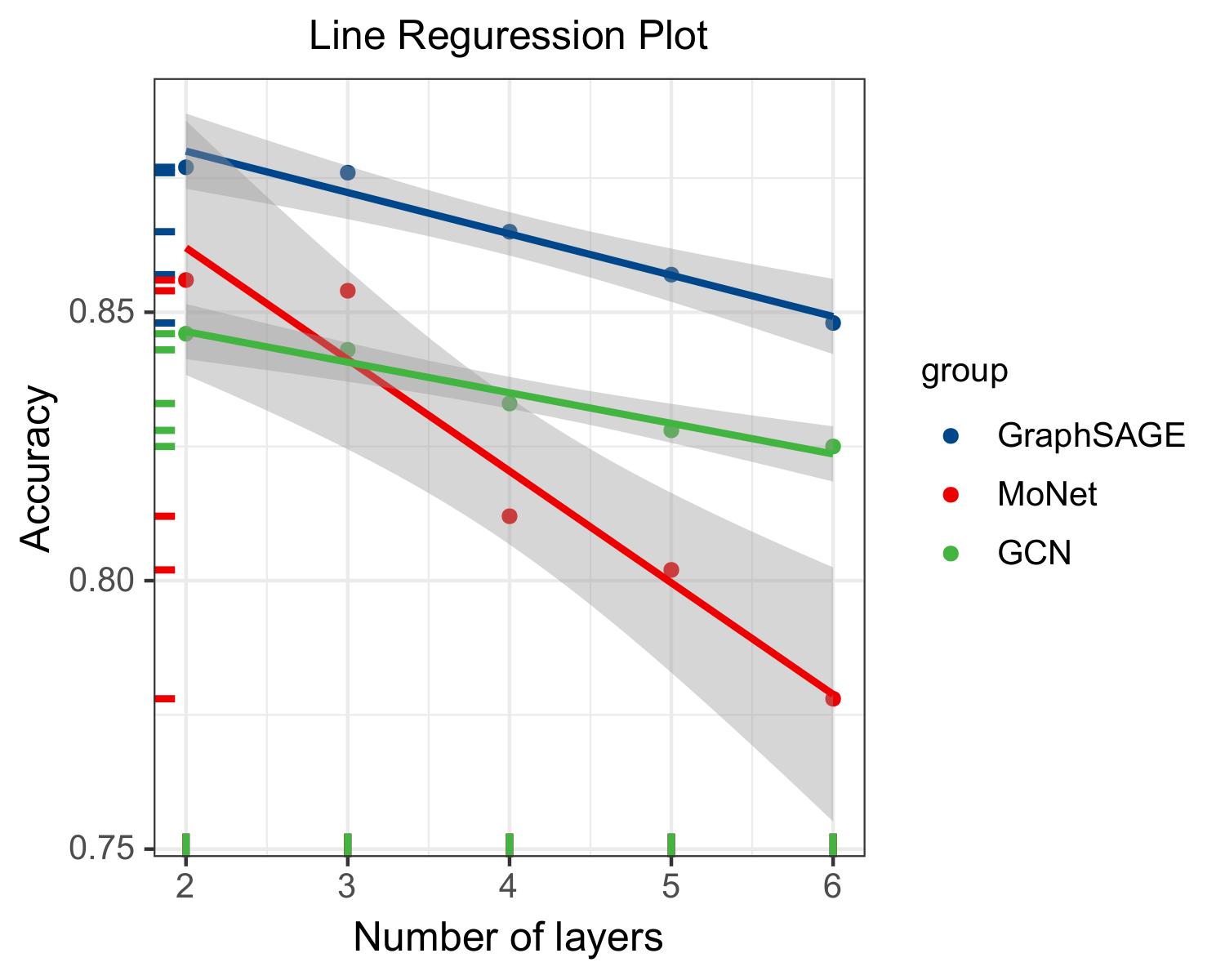}
    \caption{The existence of over-smoothing in graph neural networks}
\end{figure}

From the figure (\ref{fig:oversmoothing}), it can be seen that in a variety of GCNs implemented based on the message passing mechanism, the prediction performance reaches saturation and declines as the number of layers increases.Therefore, it is more common to build shallow graph GCNs with only 2-3 layers. While specially designed graph GCNs, such as GCNII, address the over-smoothing problem by utilizing a residual connection structure similar to that in image domains, the indiscriminate aggregation of neighboring node information still adversely impacts model performance to a certain extent. The additional computational effort required by this approach cannot be ignored. It is crucial to find ways to improve the performance of GCNs while controlling the amount of parameters.


The MSA, which is widely used in Transformer, can calculate the similarity of two nodes' representations, and then selectively choose the information passed by neighbors for weighted reception. By selectively receiving different information through the attention mechanism, nodes are able to filter out useful information and discard noisy information, which is more conducive to the functional differentiation of nodes, i.e., making different types of nodes more distinguishable. It is beneficial for downstream node classification tasks.

However, given the unique characteristics of graph data's non-grid structure, applying self-attention mechanisms to graph neural networks requires special consideration. If applies the attention mechanism to all node pairs and aggregates the global topology, although it can increase the expressiveness of the model, the $O(n^2)$ time complexity will be a huge burden. Therefore, GKEDM chooses to trade-off the expressiveness and computational overhead by choosing to aggregate node information in the first-order neighborhood of nodes with MSA. Unlike GAT, GKEDM adopts the paradigm of Transformer multi-headed attention mechanism and adds location encoding to enhance the expressiveness and differentiation of the nodes.


\subsubsection{Method}

The core goal of GKEDM is to enhance and coalesce node attribute information and graph topology information on the backbone of various types of GCN networks that have been trained to obtain a better node representation. the algorithm flow of the knowledge enhancement part of GKEDM is shown in Figure \ref{fig:enhance_module}.

\begin{figure}[H]
    \centering
    \label{fig:enhance_module}
    \includegraphics[width=1\textwidth]{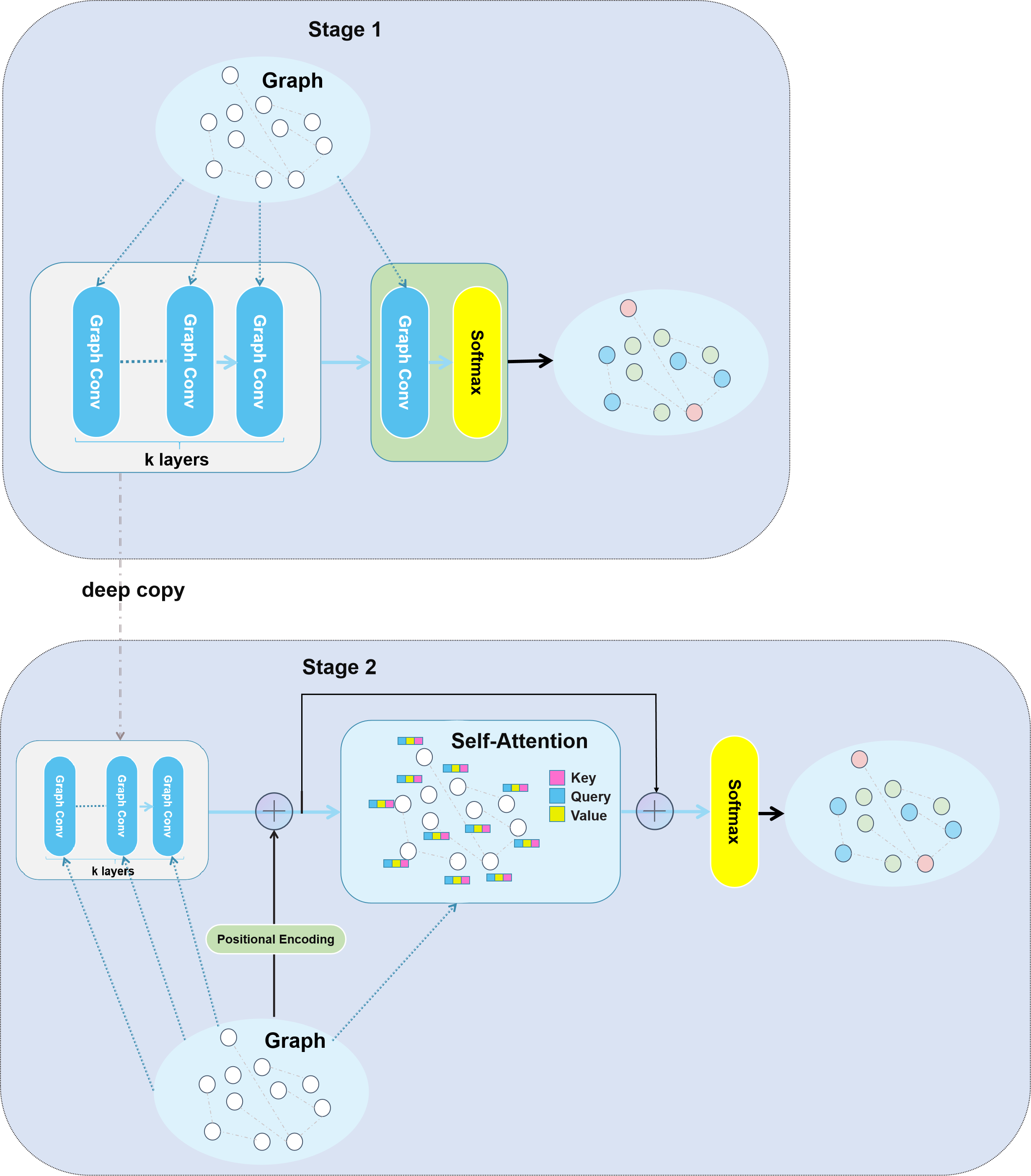}
    \caption{\textbf{GKEDM knowledge enhancement module}:GKEDM knowledge enhancement module consists of two phases. In the first phase, GKEDM trains a GCN. In the second stage, GKEDM extracts the trained GCN's backbone and concats it with the GKEDM knowledge enhancement module for fine-tuning.}
\end{figure}

GKEDM's knowledge enhancement module is a two-stage algorithm. In the first stage, GKEDM first pre-trains a GCN, as shown in the upper part of Figure \ref{fig:enhance_module}. We assume that the lower layers of GCN are mainly responsible for learning the representation of nodes, while at higher layers it is more concerned with aggregating the topology into the graph node representation. Thus the main task at this stage is to make the lower convolutional layers of the GCN learn the basic representation of a node through optimization.


After learning the basic representation of nodes, GKEDM replaces the last convolutional layer of the GCN with a MSA based layer in order to allow the network to better aggregate the information of neighboring nodes in the higher convolutional layers and thus enhance the capture of topology. The MSA based graph convolutional layer can perform more effective message aggregation.
Instead of choosing to add an additional MSA based layer after the last layer of the convolutional layer of the trained GCN, GKEDM chose to perform a replacement, an approach with two considerations:
\begin{itemize}
\item The excessive stacking of graph convolutional layers may cause information loss and smoothing of node features. Therefore, reducing the number of layers of graph convolutional neural network can avoid the over-smoothing phenomenon.
\item Reducing the number of layers of graph convolution layers can reduce the computational burden to some extent.
\end{itemize}

To apply the GKEDM module on the GCN, for a pre-trained GCN with $k+1$ layers, a node feature matrix $H_{k} \in \mathbb{R}^{n \times d}$ is generated at the $k$th layer after feeding graph $G$. At this point $H_{k}$ can be considered to have learned the basic representation of each node. To further differentiate the nodes and enhance their distinguishability, thus facilitating the downstream node classification task, GKEDM computes a positional encoding (PE) for each node. PE has proven to be very effective for applications in Transformer in the image and text domains. The principle of its action is to allow the network to distinguish the positional information of the input. In graph, nodes may have different responsibilities, communities, etc., and PE can provide the network with effective priori by using this information. GKEDM uses the Laplace position code \cite{Dwivedi_Joshi_Laurent_Bengio_Bresson_2020}. The Laplacian matrix is a very important mathematical tool to measure the graph properties and it can represent the location characteristics of the nodes in the graph. Laplacian PE is done by extracting the top $m$ smallest non-repeating eigenvalues in the graph Laplacian matrix as the eigenvectors for location coding. After combining the above position encoding, the new input $\hat{H_{k}}$ of GKEDM is computed as Eq. (\ref{con:equ6}):
\begin{equation}
    \begin{aligned}
        \hat{H_{k}} = H_{k} + f(PE(G))
    \end{aligned} \label{con:equ6}
\end{equation}

$PE$ is the Laplace PE, and $f$ is a linear mapping layer that maps the PE to the $d$-dimension node hidden vector. After obtaining $\hat{H_{k}}$, the MSA layer is computed using Eq. (\ref{con:equ3}) and Eq. (\ref{con:equ4}) to get $Attention$, and residual connection is used to stabilize the performance.
\begin{equation}
    \begin{aligned}
        K &= f_k(\hat{H_{k}}), \\
        Q &= f_q(\hat{H_{k}}), \\
        V &= f_v(\hat{H_{k}}), \\
        A &= Softmax(QK^T/\sqrt{d}) \\
        H_{k+1} &= AV + \hat{H_{k}},
    \end{aligned} \label{con:equ11}
\end{equation}

Finally, the GKEDM can be optimized by the loss function of Eq. (\ref{con:equ2}) and the stochastic gradient descent algorithm.

\subsection{Graph knowledge distillation module}
\subsubsection{Reason}
Although the knowledge enhancement module of GKEDM is able to improve the performance of the GCN through MSA, it introduces additional parameters proportional to the length $d$ of the hidden vector of the node representations due to the computation of $Q,K,V$. When $d$ is large, adding GKEDM for knowledge enhancement introduces a large computational burden.

Knowledge distillation is a common model compression technique applied in neural networks. Since the classification performance of GCN has been greatly improved by GKEDM, knowledge distillation can be performed at the cost of a small performance loss, and a trade-off between computational speed and accuracy can be made to achieve both improved accuracy and reduced model size.

\subsubsection{Method}
There's some research on how to apply knowledge distillation to graph neural networks \cite{Yang_Qiu_Song_Tao_Wang_2020}\cite{He_Wang_Zhang_Wu_2022}\cite{Joshi_Liu_Xun_Lin_Foo_2022}. In this work, a customized distillation method for GKEDM, named \emph{Attention Map Distillation (AMD}), is used, which is different from the above method. AMD is closer in principle to \cite{Yang_Qiu_Song_Tao_Wang_2020}'s Local Structure Perserving (LSP) structure, so we first give a brief introduction to the LSP structure.

The LSP structure treats the 1-hop neighborhood of a node as distribution, and the purpose of its distillation is to train the student network to mimic the 1-hop neighborhood distribution learned by the teacher network. To be more precise, as GCN goes deeper and incorporates topological information into node representations, the distribution of node neighborhoods can be illustrated as the similarity between neighboring node representations and central node representations in the hidden space. In the LSP, the similarity between nodes is calculated with three kernel functions:
\begin{equation}
    \begin{aligned}
        K(h_n^i, h_n^j)=\left\{
                        \begin{aligned}
                        & ((h_n^i)^{\top}h_n^j)^d, & Poly \\
                        & e^{-\frac{1}{2\sigma^2}}||h_n^i - h_n^j||^2 & RBF\\
                        & (h_n^i)^{\top}h_n^j,  & Linear
                        \end{aligned}
                        \right.
    \end{aligned} \label{con:equ7}
\end{equation}
The kernel function can map the low-dimensional features into the high-dimensional feature space, thus making the samples that can not be separated in the low-dimensional space linearly separable in the high-dimensional space. Among the above three kernel functions, the RBF kernel function is the most widely used, which can map the samples into an infinite dimensional feature space with good nonlinear fitting ability and adaptability.
LSP uses the KL divergence to measure the difference in the 1-hop neighborhood distribution around each node for teachers and students, and uses this as a loss function for optimization:
\begin{equation}
    \begin{aligned}
        L_{LSP} = \sum_{i \in \mathcal{V}}D_{KL}(\mathop{softmax}_{i,j \in E}(K((h_n^i)^S, (h_n^j)^S))||\mathop{softmax}_{i,j \in E}(K((h_n^i)^T, (h_n^j)^T)))
    \end{aligned} \label{con:equ9}
\end{equation}

It can be seen that the LSP uses a predefined kernel function to measure the similarity of two nodes. It may have some limitations:
\begin{itemize}
\item The assumption of using kernel functions is that the representations of two similar nodes in the kernel function space are also similar, which is not guaranteed. Similar node representations do not necessarily remain similar after the mapping of kernel functions. The kernel function approach lacks flexibility. 
\item The teacher network and the student network may have difference in parametric size, so the student network may limit its own exploration by completely imitating the structure of the teacher network in the kernel function space.
\end{itemize}

To this end, in comparison to the LSP structure that uses a predefined unlearnable kernel function to characterize the similarity between representations, the AMD proposed in this paper uses learnable parameters to measure the distance between nodes. AMD improves the effectiveness of knowledge distillation by training to fit a more suitable measure of node similarity. 
A similar approach to attention graph distillation was proposed in \cite{Wang_Wei_Dong_Bao_Yang_Zhou_2020}, but its role is in the field of natural language processing, while this paper is the first work to apply attention distillation to graph neural network distillation.

The knowledge distillation module of GKEDM is shown in the figure:
\begin{figure}[H]
    \centering
    \label{fig:enhance_module}
    \includegraphics[width=1\textwidth]{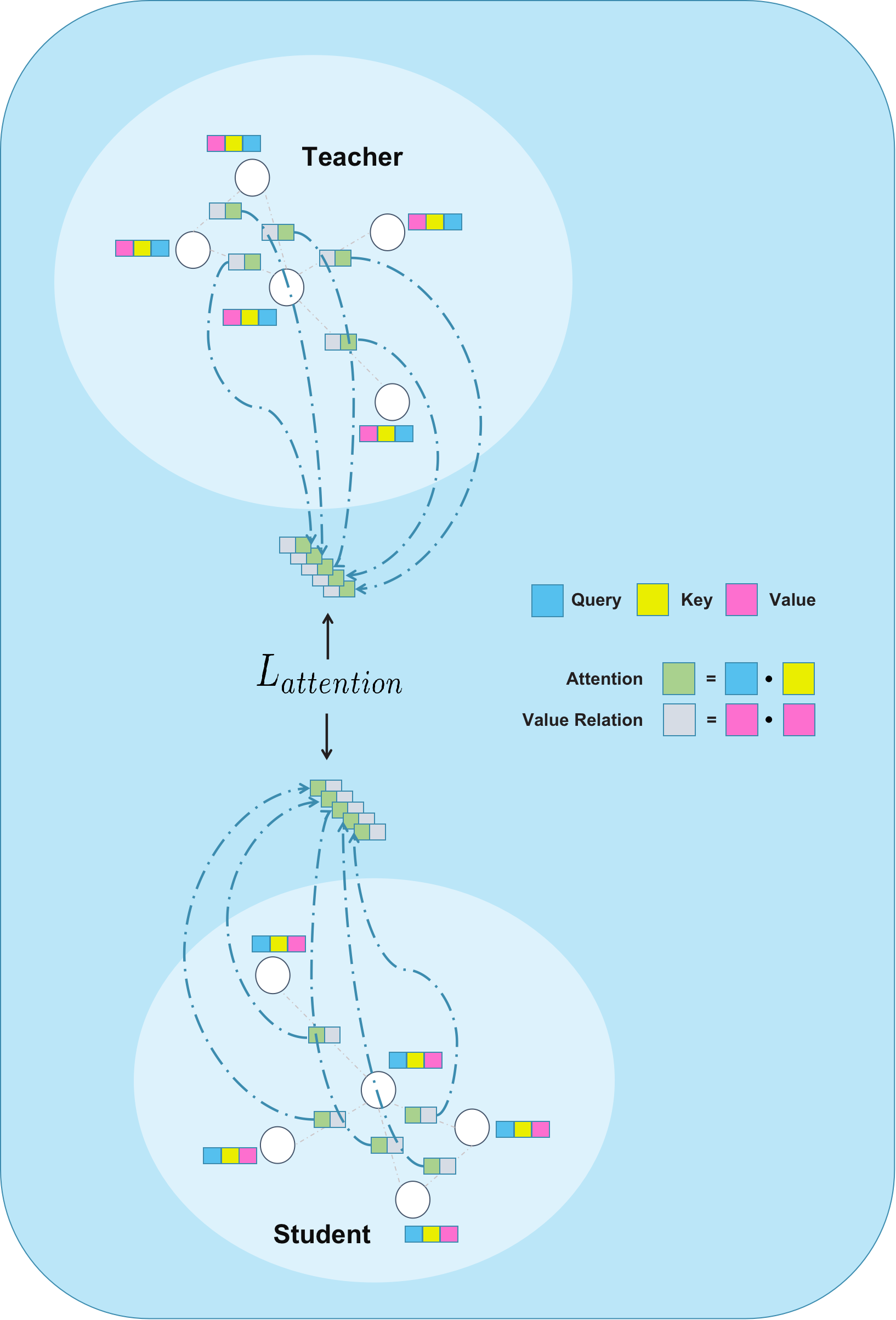}
    \caption{\textbf{Knowledge distillation module of GKEDM}:The GKEDM knowledge distillation module will drive the student network to mimic the topology of the teacher network.}
\end{figure}

The GCN for knowledge enhancement using GKEDM contains a MSA layers introduced by GKEDM. and the attention map generated by its MSA layer can be used as additional supervised information for knowledge distillation on the student network after the teacher network has completed training. The attention maps $A^T$ and $A^S$ of teacher and student respectively can be obtained by the equation (\ref{con:equ6}), which are representations of the similarity of nodes to their neighbors, and thus a description of the node neighborhood structure. By driving the student network to mimic the neighborhood structure of the teacher network, GKEDM is able to achieve the purpose of attention distillation. Different from LSP, the attention map of this method is obtained by $Query$, $key$ computation containing learnable parameters, and gains more flexibility through the MSA. This approach can overcome the distillation instability problem of teacher and student networks due to size mismatch. In particular, we train the teacher and student attention graphs by minimizing the KL divergence of:
\begin{equation}
    \begin{aligned}
        L_{A} = L_{KL}(A^T||A^S)
    \end{aligned} \label{con:equ10}
\end{equation}

With attention graph distillation, $Key$ and $Query$ pair information between teacher network nodes is transferred to students. However, in addition to Key and Query information, $Value$ are also very important representations of node descriptions in the MSA. In order to reach a more efficient distillation, Value-Value relations between teacher node pairs are also delivered as an additional attentional information for students to learn:
\begin{equation}
    \begin{aligned}
        L_{VR} = L_{KL}((V^T)^{\top}V^T||(V^S)^{\top}V^S)
    \end{aligned} \label{con:equ11}
\end{equation}

$V^T$ and $V^S$ are derived from the teacher and student, respectively. The above equation measures the dissimilarity between node pairs by the $Value-Value$ pairs. It is worth noting that this approach can be applied to Key-Key pairs and Query-Query pairs, which will be compared in the experimental section of this paper. The total loss function of the attention distillation is finally expressed as:
\begin{equation}
    \begin{aligned}
        L_{attention} = L_{A} + L_{VR}
    \end{aligned} \label{con:equ12}
\end{equation}
And the final loss function of the student network training is the cross-entropy loss plus the attention loss, which is expressed by the Eq. \ref{con:equ13}
\begin{equation}
    \begin{aligned}
        L = L_{CE} + \alpha L_{attention}
    \end{aligned} \label{con:equ13}
\end{equation}

$\alpha$ is the weight of the attention loss function, which indicates the intensity of attention distillation.

\newpage
\section{Experiments}
GKEDM is proposed to enhance the GCN for node knowledge extraction and to compress GCN's knowledge to a smaller network by distillation. In order to verify the feasibility and generality of GKEDM, we set up the experiments to answer the following questions.\\
\textbf{RQ1}:Can GKEDM work for different types of GCN on datasets of different sizes and tasks?\\
\textbf{RQ2}:Does GKEDM's improvement in performance come at the cost of additional parameters?\\
\textbf{RQ3}:Does GKEDM's attention map distillation really work?\\
\textbf{RQ4}:How does GKEDM's attention map distillation compare to other distillation methods?\\
In this section, we first introduce the dataset, the GCN and the experimental environment used, followed by experiments to answer each of the above four questions.
\subsection{Experimental setup}

\subsubsection{Dataset}
We conducted experiments on a series of datasets containing different numbers of graphs and nodes for the node classification task. These datasets include both multi-class tasks and multi-label tasks. The statistical information of the datasets is summarized in the table (\ref{tab:dataset}). More detailed information is as follows .
\begin{itemize}
\item PPI(Protein-Protein Interaction) \cite{Zitnik_Leskovec_2017} is a protein interaction dataset. Each protein is represented as a node and the interactions are represented as edges. Its contains a total of 24 graphs and is a multi-label classification task.
\item The task of the FLICKR \cite{Zeng_Zhou_Srivastava_Kannan_Prasanna_2019} dataset is to classify images on the web by their description and properties.
\item CORA FULL \cite{shchur2018pitfalls} is an expanded CORA dataset. It contains a graph representing paper citation relationships, with each node representing a paper. The task of this dataset is to predict the kind of papers.
\end{itemize}

\begin{table}[H]
\centering
\begin{tabular}{c|l|l|l}
Dataset & Number of nodes & Number of node's feature & Task \\\hline
PPI & 2,372 & 50 & Multi-label(121 labels) \\\hline
FLICKR & 89,250& 500 & Multi-class(7 classes) \\\hline
CORA FULL  & 19,793& 8710 & Multi-class(70 classes) \\\hline
\end{tabular}
\caption{\label{tab:dataset}Dataset statistics}
\end{table}

\subsubsection{Types of GCNs}
In addition to experiments on different datasets, to demonstrate that GKEDM does not work only for specific GCN, we also conducted experiments on different GCNs:
\begin{itemize}
\item GCN \cite {Kipf_Welling_2016} is the most classical semi-supervised graph neural network model. It learns node representations by defining convolutional operations on graph data structures.
\item GCNII \cite{Chen_Wei_Huang_Ding_Li_2020} avoids the over-smoothing problem existing in GCNs by residual connections, and thus can be stacked to higher layers.
\item GRAPHSAGE \cite{Hamilton_Ying_Leskovec_2017} is a GCN that learns node representations by sampling and aggregating the neighbors of nodes.
\item MONET \cite{Monti_Boscaini_Masci_Rodola_Svoboda_Bronstein_2017} is a network that can be used on non-Eulerian domains, and is therefore suitable for the graph. It enhances graph networks by using kernel functions.
\end{itemize}

\subsubsection{Experiment environment}
All experiments were performed on a LINUX system using pytorch framework version 1.12.1+cu116. The GNN framework is DGL, version 1.0.2+cu116, and the hardware used is a RTX3090.

\subsection{The enhancement effect of GKEDM for GCN (RQ1)}
To improve the performance of GCNs on node classification tasks, this work introduces an enhancement module, GKEDM. GKEDM aims at weighting the aggregated node neighborhood information and updating the basic representation of nodes by introducing an attention mechanism. However, datasets of different sizes and tasks may differ in the structure of the graph, and the semantics of the node features. Also, since the principles of different kinds of GCNs are different, the node representations they learn may also differ.
To verify whether the attention mechanism can ignore these differences and achieve its original purpose of aggregating valid information, this experiment uses different GCNs trained on different datasets and compares the classification accuracy of the network without GKEDM with that of the network with GKEDM.
Since the PPI dataset is a multi-label dataset, the metric used is the F1-score.
The experimental results are shown in the table (\ref{tab:GCNII_enhance})(\ref{tab:MoNet_enhance})(\ref{tab:GRAPHSAGE_enhance}):

\begin{table}[H]
\centering
\begin{tabular}{c|l|l|l|l|l|l}
Model &Dataset &Layers &Params(M)& Original Acc. &Acc. with GKEDM &Impv.\\\hline
GCNII & PPI & 16 &2.14 & 0.82 &\textbf{0.97 }&0.15\\\hline
GCNII & PPI & 16 &0.54 & 0.55 &\textbf{0.85} &0.30\\\hline
GCNII & CORA FULL & 1 &0.28 & 0.64 &\textbf{0.85} &0.21\\\hline
GCNII & CORA FULL & 8 &3.3 & 0.70 &\textbf{0.89} &0.19\\\hline

\end{tabular}
\caption{\label{tab:GCNII_enhance}:The effect of GCNII after GKEDM enhancement}
\end{table}

\begin{table}[H]
\centering
\begin{tabular}{c|l|l|l|l|l|l}
Model &Dataset &Layers &Params(M)& Original Acc. &Acc. with GKEDM &Impv.\\\hline
MoNet & PPI & 3 &0.03 & 0.80 &\textbf{0.93 } &0.13\\\hline
MoNet & PPI & 3 &0.54 & 0.82 &\textbf{0.97} &0.15\\\hline
MoNet & CORA FULL & 3 &5.7 & 0.65 &\textbf{0.86} &0.21\\\hline
MoNet & CORA FULL & 2 &0.28 & 0.66 &\textbf{0.81} &0.15\\\hline

\end{tabular}
\caption{\label{tab:MoNet_enhance}:The effect of MoNet after GKEDM enhancement}
\end{table}

\begin{table}[h]
\centering
\begin{tabular}{c|l|l|l|l|l|l}
Model &Dataset &Layers &Params(M)& Original Acc. &Acc. with GKEDM &Impv.\\\hline
GRAPHSAGE & PPI & 5 &0.24 & 0.76 &\textbf{0.92}&0.16\\\hline
GRAPHSAGE & PPI & 3 &0.03 & 0.66 &\textbf{0.85} &0.19\\\hline
GRAPHSAGE & CORA FULL & 2 &0.14 & 0.69 &\textbf{0.84} &0.15\\\hline
GRAPHSAGE & CORA FULL & 2 &0.28 & 0.65 &\textbf{0.86} &0.21\\\hline

\end{tabular}
\caption{\label{tab:GRAPHSAGE_enhance}:The effect of GRAPHSAGE after GKEDM enhancement}
\end{table}
From the experimental results, it can be seen that the performance of the GCNs on the node classification tasks is significantly enhanced after applied GKEDM, and the accuracy improvement of up to 30\% can be achieved. The enhancement effect is demonstrated across various GCNs and datasets, highlighting the generality of GKEDM for GCN enhancement.
It is worth noting that, as shown in Table \ref{tab:GCNII_enhance} for experiments on GCNII, different degrees of performance enhancement are observed even for same models with large differences in the number of layers and parametric quantities. It indicating that the GKEDM is a general module that works effectively for features of different scales and semantics.

\subsection{Analysis of additional parameters GKEDM introduced (RQ2)}
For neural networks, increasing the number of parameters can improve the fitting ability of the model, which is one of the most direct ways to improve the performance. The additional trainable parameters allow the model to improve the learning and to capture the information that could not be captured originally.
The knowledge enhancement module of GKEDM uses an MSA layer to replace the last layer of the original GCN. The attention layer maps node representations to $Query$, $Key$, $Value$, thus introducing additional parameters.
Although the GKEDM-enhanced model has better performance, the introduction of additional parameters makes the number of parameters more significant compared to the original model. Therefore, it is not convincing to compare only the accuracy of the original model with that of the GKEDM enhanced model.
The purpose of this experiment is to demonstrate that the improvement of GKEDM for model performance does not rely on the introduction of additional parameters, but on more efficient parameter utilization. This experiment uses the number of parameters and accuracy of the model as metrics for comparison, and the experimental results are shown in Table \ref{tab:compare}:

\begin{table}[H]
\centering
\begin{tabular}{c|l|l|l|l|l|l}
Model &Dataset &Total Param.(M) &With GKEDM &Acc.\\\hline
GCNII & PPI & 2.14 &No &0.82\\
GCNII & PPI & 0.66 &Yes &\textbf{0.85}\\\hline
GCNII & CORA FULL & 3.3 &No &0.64\\
GCNII & CORA FULL & 0.28 &Yes &\textbf{0.85}\\\hline
GRAPHSAGE & PPI & 0.24 &No &0.76\\
GRAPHSAGE & PPI & 0.14 &Yes &\textbf{0.85}\\\hline
GRAPHSAGE & CORA FULL & 0.29 &No &0.65\\
GRAPHSAGE & CORA FULL & 0.14 &Yes &\textbf{0.84}\\\hline
MoNet & PPI & 0.54 &No &0.82\\
MoNet & PPI & 0.22 &Yes &\textbf{0.93}\\\hline
MoNet & CORA FULL & 5.7 &No &0.65\\
MoNet & CORA FULL & 0.28 &Yes &\textbf{0.84}\\\hline

\end{tabular}
\caption{\label{tab:compare}:Comparison of number of Param. and Acc.}
\end{table}

As illustrated in Table \ref{tab:compare}, we performed a comparison of performance and number of parameters for the same GCN with or without the addition of GKEDM.
From the results, it can be seen that the GCN with GKEDM is able to surpass the performance of the original GCN with a small number of parameters, which proves that GKEDM does not rely on number of parameters to improve the expressive power of the model.
From the comparison of MoNet on the dataset CORA FULL below Table \ref{tab:compare}, even a small model with only one twentieth of the number of parameters can outperform a large model in the final performance through the enhancement of GKEDM.
This experiment illustrates the ability of GKEDM to improve the parameter utilization of GCNs and the effectiveness of the GKEDM for GCNs enhancement.

\subsection{GKEDM Attention Map Distillation (RQ3)}
The main purpose of this experiment is to verify the effectiveness of GKEDM attention map distillation. The task of knowledge distillation is mainly to improve the performance of the small student model by passing the dark knowledge of the large teacher model as an additional supervisory signal. Therefore, the number of teacher network participants should be more than the student network and the performance should be appropriately separated from the student network.
Therefore, this experiment was chosen to be conducted on FLICKER and PPI where the distillation effect could be seen. Meanwhile, in order to try the effect of different combinations of attentional collation, we also tried Key-Relation and Query-Relation in addition to distillation of Value-Relation. The experimental results are shown in the following figure:

\begin{table}[H]
\centering
\begin{tabular}{c|l|l|l|l|l|l|l}
Model &Dataset &\thead{Teacher\\ Param.(M)} &Teacher Acc. &\thead{Student \\ Param.(M)} &\thead{Student\\ original Acc.} &Distill Acc. &Type\\\hline
GraphSAGE & PPI & 0.74 &0.9452 &0.15 &0.8624 &\textbf{0.8755} &a+v \\
GraphSAGE & PPI & 0.74 &0.9452 &0.15 &0.8624 &0.8750 &a+v+q+k\\\hline
GraphSAGE & FLICKR & 0.4 &0.8183 &0.2 &0.6983&\textbf{0.7196} &a+v\\
GraphSAGE & FLICKR & 0.4 &0.8183 &0.2 &0.6983 &0.7147 &a+v+q+k\\\hline
MoNet & PPI & 1.04 &0.9753 &0.33 &0.9264 &\textbf{0.9399} &a+v\\
MoNet & PPI & 1.04 &0.9753 &0.33 &0.9264 &\textbf{0.9399} &a+v+q+k\\\hline
MoNet & PPI & 1.04 &0.9753 &0.21 &0.9023 &0.9281 &a+v\\
MoNet & PPI & 1.04 &0.9753 &0.21 &0.9023 &\textbf{0.9302} &a+v+q+k\\
\end{tabular}
\caption{\label{tab:KD}:Knowledge distillation effect}
\end{table}

The column $Type$ in table \ref{tab:KD} indicates the type of attention distillation used. $a,v,q,k$ represent the attention map, Value-Value relationship, Query-Query relationship and Key-Key relationship respectively. From the experimental results, it can be seen that the student network performance is improved to some extent by adding attention map distillation. There was no significant difference between distillation using only attention map and value relations and distillation using attention graphs and all relations simultaneously. The above results illustrate the effectiveness of attention distillation.

\begin{figure}[H]
    \centering
    \label{fig:weight}
    \includegraphics[width=1\textwidth]{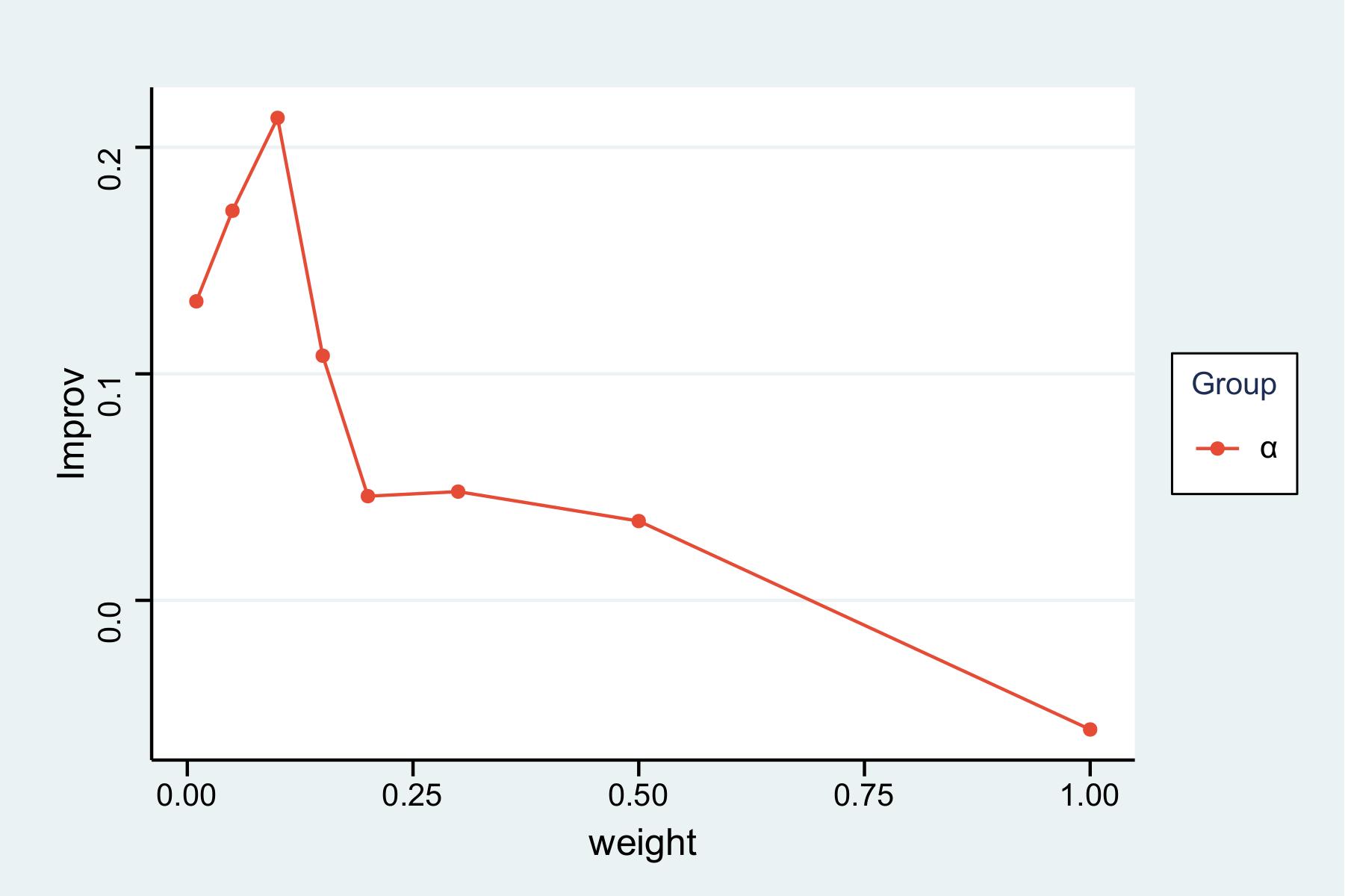}
    \caption{\textbf{Relationship between $\alpha$ setting and distillation effect}}
\end{figure}

To further investigate the effect of the parameter setting of $\alpha$ in Eq. \ref{con:equ13} on the distillation effect, we set up a control experiment of attentional distillation with different $\alpha$. The experiments use GRAPHSAGE, and the distillation method uses the attention map and Value relations as additional supervised signals. The experimental results are shown in Table \ref{tab:weight} and Figure \ref{fig:weight}.

\begin{table}[H]
\centering
\begin{tabular}{c|l}
$\alpha$ weight &Distillation Imprv.\\\hline

0.01 & +0.132\\
0.05 & +0.172\\
0.1 & \textbf{+0.213}\\
0.15 & +0.108\\
0.2 & +0.046\\
0.3 & +0.048\\
0.5 & +0.035\\
1 & -0.057\\
\end{tabular}
\caption{\label{tab:weight}:Relationship between $\alpha$ setting and distillation effect}
\end{table}

It can be observed that the attention distillation effect increases with increasing weight at the beginning and reaches its highest value at $\alpha=0.1$.  Subsequently, as the $\alpha=0.1$ increase, the distillation effect begins to decline, even impairs the performance of student network. This demonstrates that attention map distillation is better to be served as an additional supervisory signal. By constraining the student network to emulate the local topology of the teacher network during training, it allows the student network to learn to better aggregate information from topology.
When this constraint is small, the student network can learn the appropriate node representation according to the scale of its own features. When this constraint is large, it has a negative impact on the learning of the student network, where the student nodes are not free to explore the parameter space and are forced to align to the distribution of the teacher network. However, there is a scale inconsistency problem between the student model and the teacher model. Forcing the student model to align to teacher model can hurt the performance of the student model.
%
\subsection{Comparison of GKEDM distillation and related work (RQ4)}
In order to further demonstrate the effectiveness of GKEDM distillation, this experiment compares the effectiveness of distillation using GKEDM with that of related distillation methods. Other distillation methods specifically compared are:
\begin{itemize}
\item \textbf{KD}:It is the first method used to improve the performance of small models by distilling the dark knowledge of a large teacher model. The core idea is to use the output probability distribution of the teacher model as a soft label to guide the learning process of the student model, thus improving the generalization performance of the student model.
\item \textbf{FITNET}:FITNET uses an intermediate layer distillation approach. In the distillation process, the intermediate layer output of the teacher model is used as the target of the corresponding intermediate layer of the student model, and the student model learns the knowledge of the teacher model by minimizing the distance between its intermediate layer output and the intermediate layer output of the teacher model. When the middle layer feature size of the student network is different from that of the teacher network, the student network output size and the teacher network output size can be made to align by a linear mapping method.
\item \textbf{LSP}:It is the first method to apply knowledge distillation to graph neural networks. Through the Local Structure Preservation (LSP) module, the teacher network is able to guide the student network to learn the topology of graph data, thus improving its performance.

\end{itemize}
The experimental results are shown in Table \ref{tab:compare_related}. In the KD method, the hyperparameters of the student network training are set as follows: soft label distillation weight is 0.8 and hard label weight is 0.2. The parameter settings of the LSP are referred  to the original paper. The weight of LSP loss is 100, and the kernel method is set as RBF.
\begin{table}[H]
\centering
\begin{tabular}{c|l|l|l|l|l|l|l}
Model &Dataset &\thead{Teacher \\ Param.(M)} &Teacher Acc. &\thead{Student \\Param. (M)} &\thead{Student\\ original Acc.} &Distill Acc. &Type\\\hline
GraphSAGE & PPI & 0.74 &0.9213 &0.15 &0.8865 &0.8527 &KD\\
GraphSAGE & PPI & 0.74 &0.9213 &0.15 &0.8865 &0.8823 &LSP\\
GraphSAGE & PPI & 0.74 &0.9213 &0.15 &0.8865 &0.8500 &FITNET\\
GraphSAGE & PPI & 0.74 &0.9213 &0.15 &0.8865 &\textbf{0.9019} &a+v+q+k(ours)\\\hline
GraphSAGE & FLICKR & 0.4 &0.8183 &0.2 &0.6983 &0.6977 &KD\\
GraphSAGE & FLICKR & 0.4 &0.8183 &0.2 &0.6983 &0.7049 &LSP\\
GraphSAGE & FLICKR & 0.4 &0.8183 &0.2 &0.6983 &0.6771 &FITNET\\
GraphSAGE & FLICKR & 0.4 &0.8183 &0.2 &0.6983 &0.\textbf{7155}&a+v(ours)\\\hline
MoNet & PPI & 1.04 &0.9753 &0.21 &0.9023 &0.9166 &KD\\
MoNet & PPI & 1.04 &0.9753 &0.21 &0.9023 &0.9297 &LSP\\
MoNet & PPI & 1.04 &0.9753 &0.21 &0.9023 &0.9282 &FITNET\\
MoNet & PPI & 1.04 &0.9753 &0.21 &0.9023 &\textbf{0.9302} &a+v+q+k(ours)\\
\end{tabular}
\caption{\label{tab:compare_related}:Comparison of GKEDM distillation and related work}
\end{table}
From the experimental results, it can be seen that attention distillation using GKEDM achieved better results than the previous method on different GCNs and different datasets, which illustrates the importance of attention for improving the performance of GCNs. Compared with the unlearnable distance measurement used in LSP, the distance measurement based on learnable parameters of GKEDM is more flexible and reduces the constraints on the student model, thus achieving better results.

\newpage
\section{Shortcomings and future work}
Although GKEDM has achieved good results on various GCNs and datasets, the following problems remain:
\begin{itemize}
\item Since GKEDM is still a GCN implemented based on the message passing mechanism, it also suffers from the problem of over-smoothing when the number of layers is too large. It has been verified that the performance degradation still occurs after stacking the attention layers of GKEDM. This problem reduces the scope of GKEDM applications. The limitation of the number of parameters limits the effectiveness of GKEDM on large datasets. The performance of GKEDM on large-scale models and datasets has not been fully tested in this work.
\item Although GKEDM distillation is effective, the GCN has to be added with GKEDM module and retrained. For large datasets, this process is very time and resource consuming. In addition, the distillation process of GKEDM is relatively complex and is not an end-to-end process.
\end{itemize}

Therefore, the future work could be:
\begin{itemize}
\item Investigate the efficacy of GKEDM on large GCNs and large datasets, then conceive a more universal framework.
\item Simplifying the distillation process, enables GKEDM to save computational time and computational resources while maintaining the distillation effect.
\end{itemize}

\newpage
\section{Conclusion}
In this work, we propose a knowledge enhancement and distillation module for GCN, called GKEDM, which aims to improve the performance of GCNs on node classification tasks and to lightweight them. The enhancement module of GKEDM enhances the ability of GCNs to learn graph node representations by introducing a convolutional layer based on an attention mechanism, thus improving their performance on node classification tasks. Experiments show that GKEDM can provide performance enhancement for different kinds of GCNs on different datasets.

To make the model more lightweight through knowledge distillation, we also propose a distillation module suitable for GKEDM. This module allows the student network to mimic the neighborhood structure of the teacher network through attention map distillation. The experimental results show that the distillation module of GKEDM achieves the best results on several commonly used node classification task datasets, proving the effectiveness of the module. We believe that the method can provide useful guidance for broader research on GCNs and can also serve as an effective solution to improve the performance of node classification tasks. We believe that the method can provide useful guidance for broader research on GCNs and can also serve as an effective solution to improve the performance of GCNs on node classification tasks. we hope this work can stimulate more research on GCNs and provide support for broader applications in the future.


\bibliographystyle{IEEEtran}
\bibliography{sample}

\end{document}